\documentclass[10pt, conference, compsocconf, letterpaper]{IEEEtran}
\IEEEoverridecommandlockouts
\usepackage{cite}
\usepackage{amsmath,amssymb,amsfonts}
\usepackage[ruled,linesnumbered,algo2e]{algorithm2e}
\usepackage{algorithm}
\usepackage{algpseudocode}
\usepackage{multirow}
\usepackage{tabularx}
\usepackage{graphicx}
\usepackage{textcomp}
\usepackage{xcolor}
\usepackage{listings}
\usepackage{bm}
\usepackage{booktabs}
\usepackage{multirow}
\usepackage{graphicx}

\lstdefinestyle{yaml}{
     basicstyle=\color{blue}\footnotesize,
     rulecolor=\color{black},
     string=[s]{'}{'},
     stringstyle=\color{blue},
     comment=[l]{:},
     commentstyle=\color{black},
     morecomment=[l]{-}
 }
\usepackage{subfig}
\usepackage[belowskip=0pt,aboveskip=0pt,margin=0pt,font={small,it},labelfont={small,bf}]{caption}
\usepackage{enumitem}

\def\BibTeX{{\rm B\kern-.05em{\sc i\kern-.025em b}\kern-.08em
    T\kern-.1667em\lower.7ex\hbox{E}\kern-.125emX}}
\begin{document}

\title{\textit{CoverNav}: Cover Following Navigation Planning in Unstructured Outdoor Environment with Deep Reinforcement Learning\\

% \thanks{The work is funded by the U.S. Army grant \#W911NF2120076 and NSF Research Experiences for Undergraduates (REU) grant \#CNS-2050999.}
}

% \author{\IEEEauthorblockN{Wanying Zhu}
% \IEEEauthorblockA{\textit{Institute of Artificial Intelligence} \\
% \textit{\& School of Computing} \\
% \textit{University of Georgia}\\
% wz77105@uga.edu}
% \and

\author{\IEEEauthorblockN{Jumman Hossain}
\IEEEauthorblockA{\textit{Department of  Information Systems} \\
\textit{University Of Maryland, Baltimore County}\\
jumman.hossain@umbc.edu }
\and
\IEEEauthorblockN{Nirmalya Roy}
\IEEEauthorblockA{\textit{Department of Information Systems} \\
\textit{University Of Maryland, Baltimore County}\\
nroy@umbc.edu}
}

\author{\IEEEauthorblockN{Jumman Hossain$^1$, Abu-Zaher Faridee$^1$, Nirmalya Roy$^1$, Anjan Basak$^2$, Derrik E. Asher$^2$}
\IEEEauthorblockA{$^1$Department of Information Systems,
University of Maryland, Baltimore County, USA}
\IEEEauthorblockA{$^2$DEVCOM Army Research Lab, USA}
\IEEEauthorblockA{$^1$\{jumman.hossain, faridee1, nroy\}@umbc.edu} $^2$anjon.sunny@gmail.com, $^2$derrik.e.asher.civ@army.mil}

\maketitle

\begin{abstract}
Autonomous navigation in off-road environments has been extensively studied in the robotics field. However, navigation in covert situations where an autonomous vehicle needs to remain hidden from outside observers remains an under-explored area.
In this paper, we propose a novel Deep Reinforcement Learning (DRL) based algorithm, called \textit{CoverNav}, for identifying covert and navigable trajectories with minimal cost in off-road terrains and jungle environments in the presence of observers. \textit{CoverNav} focuses on unmanned ground vehicles seeking shelters and taking covers while safely navigating to a predefined destination. Our proposed DRL method computes a local cost map that helps distinguish which path will grant the maximal covertness while maintaining a low-cost trajectory using an elevation map generated from 3D point cloud data, the robot's pose, and directed goal information. \textit{CoverNav} helps robot agents to learn the low-elevation terrain using a reward function while penalizing it proportionately when it experiences high elevation. If an observer is spotted, \textit{CoverNav} enables the robot to select natural obstacles (e.g., rocks, houses, disabled vehicles, trees, etc.) and use them as shelters to hide behind. We evaluate \textit{CoverNav} using the Unity simulation environment and show that it guarantees dynamically feasible velocities in the terrain when fed with an elevation map generated by another DRL-based navigation algorithm. Additionally, we evaluate \textit{CoverNav's} effectiveness in achieving a maximum goal distance of 12 meters and its success rate in different elevation scenarios with and without cover objects.  We observe competitive performance comparable to state-of-the-art (SOTA) methods without compromising accuracy.
% We also implemented \textit{CoverNav} on a Husky robot and presented the maximum goal distance (12 meters) to reach and success rate in the presence of different types of elevation with or without cover.
\end{abstract}


\section{Introduction}
 
% It can be divided into several subtopics based on the environment that the robot operates in primarily, such as on-road, off-road, indoor environments, etc. 

\begin{figure}[!htb]
\centering
\includegraphics[width=0.5\textwidth]{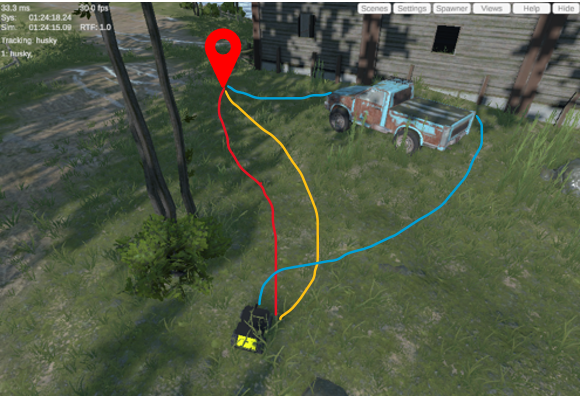}
% \caption{The Clearpath Husky robot navigating on our method in simulation. The orange path indicates the robot's path running on our DRL-based method. Pink represents TERP\cite{terp}, and blue represents GANav\cite{ganav}, both are state-of-the-art navigation planning algorithms. Our fully trained robot detects the position of its nearby obstacles, and moves towards it to find cover. In this image, the robot first finds the bush to be its cover, then once it perceives that no more cover is in the direction it is heading, it quickly moves in a different direction to find shelter between the truck and the cottage until it reaches the destination.}
\vspace{2pt}
\caption{
During navigation in unstructured outdoor environments, our proposed \textit{CoverNav} algorithm directs the Clearpath Husky robot towards nearby obstacles that provide the highest cover. In simulation, the robot detects the position of obstacles in its vicinity and moves towards them to find cover. If there are multiple obstacles in the robot's field of view, \textit{CoverNav} prioritizes the obstacle with the highest cover. In the accompanying image, we show the robot navigating using our DRL-based method in blue, and state-of-the-art navigation planning algorithms, TERP~\cite{weerakoon2022terp} in red and GA-Nav~\cite{guan2022ga} in yellow. In the image, the robot initially detects a bush as its cover, moves towards it, and then changes direction when it no longer perceives any cover in that direction. It quickly finds shelter between a truck and a cottage until it reaches its destination.
}
\label{fig}
\end{figure}

% In military-relevant environments, in order for soldiers to properly execute their objectives, they must navigate a broad array of landscapes. As human-agent teaming and artificial intelligence continue to progress, robots will need to demonstrate the same level of adaptability in order to successfully navigate different types of terrain and become effective combat companions. An autonomous vehicle has the ability to serve as a battle companion to a human soldier while working side-by-side with them or actively taking part in the battle, putting themselves in a variety of advantageous positions to conduct operations against the adversary. % A vehicle might have to make a choice between driving over grass, sand, or mud, all of which could make autonomous maneuvering more difficult.

Humans and animals can traverse challenging terrain and off-road environments with ease to reach their destination.
As the fields of human-autonomous agent teaming and artificial intelligence progress, it becomes essential for robots to showcase a similar level of adaptability, enabling them to navigate diverse terrains effectively and become valuable companions in various scenarios. Autonomous ground vehicles, particularly, have great potential as companions to humans, seamlessly collaborating with or actively participating in tasks. They can strategically position themselves in advantageous locations to optimize operations and achieve objectives more efficiently. In order for autonomous ground vehicles to be effective in various scenarios, they must be able to navigate through complex terrains. These terrains include urban environments with paved surfaces as well as off-road areas with natural obstacles like forests, jungles, deserts, and uneven terrain with wet gaps. Additionally, rural settings with boundary fences, walls, and sparse structures pose unique challenges.
% In each of these scenarios, contact with potential adversarial positions is a constant concern – in some situations, this contact should be avoided through the use of terrain features and cover. 
In each scenario, various physical features can offer sufficient terrain cover for autonomous ground vehicles to minimize their visibility. 
This allows them to increase their chances of mission success, especially in situations where it's necessary to move the robot while remaining hidden from observers. By utilizing these physical features effectively, autonomous ground vehicles can enhance their ability to remain concealed, reduce the risk of being detected, and improve the overall effectiveness of their missions. 

Al Marzouqi and Jarvis \cite{al2011robotic} conducted one of the earliest comprehensive surveys on robotic covert path planning, outlining the various strategies and techniques used in the field. Subsequently, Ninomiya et al. \cite{ninomiya2015planning} presented a planning framework that satisfies multiple spatial constraints imposed on the path, such as staying behind a building, walking along walls, or avoiding the line of sight of patrolling agents.  As evident from these works, to achieve effective covert path planning, UGVs must have an integrated navigation planning system that allows them to adapt their movements based on the environment and select routes that optimize cover and concealment.

In order to effectively utilize the terrain covers to keep unmanned ground vehicles (UGVs) hidden, proper logic must be integrated into their navigation planning system. 
% Effectively utilization of terrain cover and remaining hidden can only be realized with proper navigation planning on these unmanned ground vehicles (UGVs). 
UGVs must navigate the ground without direct supervision and adapt their movements based on the environment instead of following a fixed path. It is important for UGVs to select routes that optimize cover and concealment as if they expect to be seen by an observer at any moment. 
% This strategic approach to navigation planning ensures that UGVs minimize their visibility and significantly increase their chances of successfully accomplishing their objectives. By integrating advanced perception systems and intelligent decision-making algorithms, UGVs can plan their paths in a way that maximizes the use of terrain cover.
Developing such an effective off-road navigation planning method presents a significant challenge: \textit{distinguishing safe and traversable regions for a mobile robot while considering its properties, constraints, and terrain cover}. Many existing methods predominantly focus on environmental elements such as surface properties and often overlook the dynamic constraints of the robot, undermining the potential impacts on their interactions with the environment. Semantic segmentation-based algorithms categorize various terrain surfaces by learning their visual characteristics, such as color and texture~\cite{guan2022ga, hirose2018gonet, shaban2021semantic, viswanath2021offseg, singh2021offroadtranseg, maturana2018real,valada2017deep, dabbiru2020lidar, jiang2021rellis, wigness2019rugd}. However, these algorithms often fail to consider the dynamic constraints of the UGV, thereby neglecting the impact of these constraints on the environmental interaction. 

% The majority of DRL-based navigation techniques\cite{weerakoon2022terp,choi2019deep} are utilized for detecting unsafe elevation changes, efficient trajectory planning, or guiding a robot around moving objects. However,  current DRL approaches often do not consider guiding the robot while maintaining safe-cover in order to avoid enemies in a military-relevant environment.

% Another advancement for autonomous navigation is Self-Supervised Learning for Unstructured Terrain Navigation ~\cite{sathyamoorthy2022terrapn}. While self-supervised learning presents a compelling approach for Unmanned Ground Vehicles (UGVs)~\cite{castro2022does, kahn2021badgr,zhou2012self} to learn from their own experiences and adapt to navigate effectively, it is not without its limitations. One of the significant challenges is that self-supervised learning depends heavily on the quality and diversity of the data the robot can collect. If the robot operates in a limited environment or encounters a lack of diverse experiences, it may fail to develop a comprehensive understanding of different terrains and obstacles, leading to sub-optimal navigation strategies.  It may struggle to handle dynamic elements, such as moving obstacles, changing terrain conditions, or adversarial actions in a military context, as it learns from static snapshots of the surrounding environment. 

Deep Reinforcement Learning (DRL) is a promising approach that can potentially address some of these limitations for navigation planning~\cite{Zhang2018RobotNO,zhou2022learning} and UGV can learn to make strategic decisions based on the environment's current state to maximize a reward function. DRL techniques have excelled in detecting unsafe elevation changes, efficient trajectory planning, and guiding UGVs around moving objects~\cite{choi2019deep}.
However, they often do not account for the crucial aspect of minimizing the likelihood of being seen by an outside observer. Incorporating these strategic considerations into the DRL framework is an essential step toward creating more effective and practical UGVs for off-road navigation.
% However, they often do not account for the crucial aspect of maintaining safe cover to avoid enemies in a military-relevant environment. Incorporating these strategic considerations into the DRL framework is an essential step toward creating more effective and practical UGVs in combat scenarios.

% Most existing DRL-based navigation algorithms do not have this feature, which motivates us to design a DRL-based method to incorporate the idea of seeking maximum cover during robot navigation. Our approach plans covert and navigable trajectories in unstructured outdoor terrains and guarantees dynamically feasible velocities accounting for the robot’s dynamic constraints. Although this approach is more likely to generate longer trajectories, finding cover during navigation may be desired in contested environments like battle fields, where the robot needs to stay relatively invisible to be safe.

% \textbf{Main Contributions:} 
In this work, we present a novel Deep Reinforcement Learning (DRL) based method for identifying covert and navigable trajectories in off-road terrains and jungle environments. Our approach focuses on seeking shelters and covers while safely navigating to a predefined destination with the objective of minimizing the visibility of the robot to potential observers. 
In the context of this problem, cover refers to any physical object that can shield or conceal the robot from detection by observers. Examples of such objects include buildings, trees, hills, vehicles, and any other obstacles that can obscure the robot's view or hide it from the line of sight of an observer.\\

\noindent Our main contributions are summarized as the following:
% We present a novel Deep Reinforcement Learning (DRL) based method for identifying covert and navigable trajectories in off-road terrains and jungle environments. Our approach focuses on seeking shelters and covers while safely navigating to a predefined destination.
\begin{figure*}[!ht]
\centering
\includegraphics[trim=6 530 6 150, clip,width=\textwidth]{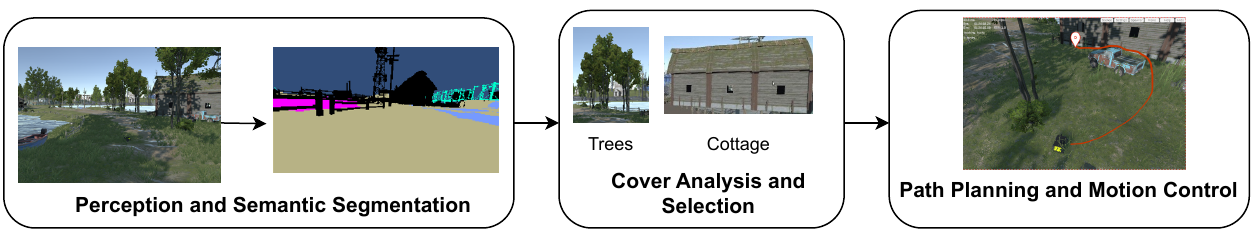}
\caption{ \textbf{Overview of the \textit{CoverNav} Navigation Planning System}: The process begins with \textbf{Perception and Semantic Segmentation}, where the environment is sensed and objects are classified (block 1). This feeds into the \textbf{Cover Analysis and Selection} module (block 2), which identifies potential cover objects and selects the optimal one. The selected cover object is then used by the \textbf{Path Planning} and \textbf{Motion Control} modules (block 3) to both generate a safe and optimal trajectory towards the object and executes this trajectory through the robot's physical movements.}
\label{coverNav}
\end{figure*}

\begin{itemize}[leftmargin=*]
%   \item We propose a DRL network that uses a novel reward function to learn appropriate off-road terrain features with covers in simulated environment. The reward designed to  have reward for going in the direction of the objective, penalty for straying from the direction of the objective, reward for maintaining a low altitude, reward for moving near to cover, and reward for maintaining stability. Our proposed approach learns fast, plans reliable and steady trajectories and maintains close but safe distance to obstacles to seek covers.

\item \textbf{Designing a Novel Reward Function :} Our proposed DRL network utilizes a novel reward function to effectively learn optimal off-road terrain features with covers in simulated environments. By rewarding the agent for moving towards the objective, penalizing for deviating from it, promoting low altitude, encouraging proximity to cover, and maintaining stability, our approach is able to learn quickly and efficiently. This results in the ability to plan reliable and steady trajectories while maintaining a safe distance from obstacles in order to seek out covers.

% \item We design a novel cover-detecting algorithm by using existing semantic segmentation approaches. The algorithm has the capability of identifying the presence and location of cover objects, and it may make use of these things to shield the robot whenever it is necessary to do so. Instead of just detecting cover objects in a bounding box, which can lead to imprecise findings, the method can precisely identify and locate cover objects with the uses of semantic segmentation. This allows for more accurate results.

\item \textbf{Designing a Novel Cover Detection Algorithm : }We present a novel cover detection algorithm that leverages state-of-the-art semantic segmentation techniques to precisely identify and locate cover objects in unstructured outdoor environments. By exploiting the rich semantic information provided by segmentation, our algorithm can accurately identify and localize cover objects, providing the necessary information to the robot to take appropriate actions for seeking shelter when needed. The proposed approach thus enhances the safety and robustness of the robot navigation system in challenging outdoor environments.

% Our method overcomes the limitations of conventional bounding box-based methods, which can produce imprecise results. 

%   \item We integrate TERP (Terrain
% Elevation-based Reliable Path Planning)\cite{terp} - a novel method for navigating a robot in an outdoor environment with reliability and stability and DWA-RL\cite{dwarl} - an approach to navigation in an environment with mobile obstacles that combines the advantages of traditional DWA and DRL-based techniques. These integration's assist us in locating cover and shelter while also facilitating the safe navigation to a predetermined destination and the avoiding of dynamic obstacles. 

\item \textbf{Integrated Cover-Following Navigation Planning Approach:} We propose an integrated approach that combines Terrain Elevation-based Reliable Path Planning (TERP)~\cite{weerakoon2022terp} and Dynamic Window Approach-based Reinforcement Learning (DWA-RL)~\cite{patel2020dynamically} with our cover detection algorithm to enhance the navigation of a robot in an unstructured outdoor environment. The TERP method provides reliable and stable path planning for the robot, while the DWA-RL approach facilitates navigation in an environment with mobile obstacles. The integration of these methods enables the robot to locate cover and shelter, navigate safely (i.e., maintain a stable trajectory and move without being detected by potential threats) to a predetermined destination, and avoid dynamic obstacles. This combined approach provides a more robust and efficient navigation strategy for the robot in complex outdoor environments.

% \item We evaluate the cover-following navigation performance in simulated unstructured environments such as off-road terrains and jungle with sparse or dense grasses using a Clearpath Husky robot. In addition, we analyze our strategy in a realistic forest environment, which includes a variety of terrain, extensive areas of cover, and an adequate range of elevations. We demonstrated the highest achievable goal distance (12 meters) and success rate when encountering varying degrees of elevation with and without cover. This helps us better understand how our plan will perform in the real world.

\item \textbf{Experimenting in Various Simulated Unstructured Environments:} In order to evaluate the effectiveness of our proposed approach, we conducted extensive simulations using a Clearpath Husky robot in various unstructured environments, such as off-road terrains and jungles with varying levels of vegetation density. CoverNav exhibits competitive performance in success rate and trajectory length, comparable to state-of-the-art (SOTA) methods while maintaining high accuracy. We also tested our strategy in a realistic forest environment with diverse terrain features, extensive areas of cover, and different elevations. Our experiments demonstrate that \emph{our approach can successfully achieve the highest possible goal distance of 12 meters and high success rates} (i.e., the percentage of trials in which the robot successfully completes the specified goal distance) in navigating through terrains with varying degrees of elevation and cover. These results provide valuable insights into the potential real-world performance of our approach.

\end{itemize}

\noindent The rest of the article is organized in the following structure. Section \ref{sec:relatedwork} covers related prior works in off-road path planning. Section \ref{sec:methodology} provides a brief overview of the architecture, the background information as well as the proposed methodology. A comprehensive analysis of the experiments and their outcomes can be found in section \ref{sec:experiment}. In the following section~\ref{sec:discussion}, there is a brief discussion about the limitations of our approach. In the final Section (\ref{sec:conclusion}), we summarize our findings and discuss potential future directions.

\section{Related Work}
\label{sec:relatedwork}

% This section discusses prior works in outdoor navigation planning methods, including a semantic segmentation approach, a self-supervised learning approach, and a DRL approach.

In this section, we discuss existing work on robot navigation in unstructured outdoor environments.

\subsection{Semantic Segmentation for Off-Road Surface Traversability}

In many scenarios, a robot encounters unstructured outdoor terrains with different levels of navigability. Many of these terrain classes have extremely similar visual features and overlaps that can lead to misclassification and result in dangerous navigation. Several recent works have addressed these challenges of off-road navigation by proposing novel methods with semantic segmentation ~\cite{gao2021fine, shaban2022semantic}. GA-Nav~\cite{guan2022ga} is a learning-based semantic segmentation method that identifies different terrain groups in off-road and unstructured outdoor terrains. It balances segmentation accuracy and computational overhead and outperforms state-of-the-art segmentation methods by using coarse-grained segmentation. For instance, it recognizes trees and poles simply as obstacles instead of categorizing them as individual classes. In the context of our work, we utilize the potential of the semantic segmentation method to effectively detect and classify cover objects in the environment.

% \subsection{Self-Supervised Learning for Unstructured Terrain Navigation}
% Unsupervised learning-based methods have been also investigated in robot navigation as they do not require human annotated datasets for training. Rather, these methods stress on automating the labeling process by collecting data from the robot’s interaction with different terrain surfaces and associating them with various visual features (RGB images). The terrain-interaction includes many aspects including torques, contact vibrations, acoustic data, vertical acceleration, and stereo depth. 
% TerraPN\cite{sathyamoorthy2022terrapn} uses online self-supervised learning to predict a surface cost map and uses it for efficient robot navigation in outdoor terrains. During the training process, an autonomous robot collects RGB images and odometry data by traveling through various surfaces with changing velocities. This method combines the benefits of accurate characterization of robot-terrain interaction and provides low surface cost and dynamically feasible navigation. Another recent study by Guaman Castro et al.\cite{castro2022does} focuses on estimating terrain traversability in off-road environments through self-supervised costmap learning. Their method combines exteroceptive environmental information with proprioceptive terrain interaction feedback to predict traversability costmaps. The incorporation of robot velocity in the costmap prediction pipeline enhances the fine-grained understanding of robot-terrain interactions. 

\subsection{Deep Reinforcement Learning Based Off-road Navigation}

% Terrain elevation data are necessary to consider if a trajectory planning algorithm strives for robust outdoor navigation. Some elevation changes may be dangerous for specific mobile robots due to their dynamic and physical constraints. Steep slopes and other forms of sudden elevation changes need to be avoided by the robot to ensure stable and safe trajectory. Previous studies ~\cite{fankhauser2014robot} gather elevation information using camera or LiDAR sensors as the robot moves, and the information is perceived using grid-based data structures like Octomaps ~\cite{hornung2013octomap} and elevation maps.

% TERP~\cite{weerakoon2022terp} is a DRL-based method for navigating a robot in a reliable and stable manner in uneven outdoor terrains. It identifies unsafe regions, which are characterized by high elevation gradients, and computes local least-cost waypoints to avoid those untraversable regions.

% However, none of these existing methods \cite{wiberg2021control,josef2020deep,zhou2022learning, Zhang2018RobotNO, triest2023learning} address the feature of finding places to hide in outdoor terrains, which motivates the current project. The proposed method compares with some approaches to evaluate its performance.

% Previous studies have tackled the issue of gathering elevation information, typically employing cameras or LiDAR sensors as the robot moves, and interpreting this information using grid-based data structures such as Octomaps and elevation maps \cite{fankhauser2014robot, hornung2013octomap}.

In recent studies, Deep Reinforcement Learning (DRL) has been utilized to improve navigation strategies in complex, off-road environments. Taking the challenge of uncertainty, Zhang et al.~\cite{Zhang2018RobotNO} demonstrated the potential of DRL for navigating environments with unknown rough terrain. Wiberg et al.~\cite{wiberg2021control} proposed a DRL approach for vehicle control in rough terrains, concentrating on developing an effective mechanism to handle complex and uneven surfaces. Similarly, Josef and Degani~\cite{josef2020deep} presented a DRL method for safe local planning of a ground vehicle in unknown rough terrain, underscoring the necessity for robust navigation in the face of  unpredictability and hazards associated with rough terrains.  Further expanding the DRL framework, Zhou et al.~\cite{zhou2022learning} introduced a multimodal approach to navigating rough terrain, reinforcing the efficacy of integrating multiple strategies within a DRL framework for enhanced navigation robustness. Another recent DRL-based method, TERP~\cite{weerakoon2022terp} is specifically designed to facilitate robust navigation in uneven outdoor terrains. TERP excels in identifying unsafe regions characterized by high elevation gradients and in calculating local least-cost waypoints to circumvent the untraversable areas.

While existing techniques have achieved considerable advancements in autonomous navigation, one important aspect that has been overlooked is the ability to find places for hiding or taking cover in outdoor terrains. This crucial feature is particularly relevant in military scenarios, where maintaining concealment from potential observers is paramount. In our method, we propose a novel approach to incorporate the feature of cover-seeking into the DRL-based navigation of unmanned ground vehicles in off-road terrains. Our approach represents a promising step towards enabling more robust, and intelligent navigation in complex environments.

% \section{Overview}

% Our hypothesis is that maintaining a close distance to nearby obstacles like trees while avoiding collisions is the gist to safely find maximum cover in unstructured outdoor environments. We also think that avoiding elevation gain will help the robot stay unnoticed.

% However, if no shelters are present near the robot, or if nearby shelters are in unsafe regions, the robot should navigate directly to the predefined location by planning a short and safe trajectory. Then, it follows the path with dynamically feasible velocities. Our approach adapts and integrates the following existing algorithms for the perception of the environment and the navigation planning model.

% \begin{figure}[htbp]
% \centerline{
% \includegraphics[width=0.50\textwidth]{images/system architecture.png}}
% \caption{Our system architecture.}
% \label{fig}
% \end{figure}

% This method gathers raw sensor data from the simulation environment, and parse them to pose, twist and elevation map as the inputs. Then the inputs enter a perception module, which is a DRL network based on the Deep Deterministic Policy Gradient (DDPG) algorithm combined with Convolutional Neural Network Layers. The DRL network then outputs an elevation cost map to the Dynamic Window Approach (DWA) to generate Dynamically feasible velocities.

\section{Methodology}
\label{sec:methodology}
%=========================================================================
In this section, we detail the building blocks of \textit{CoverNav}; we first start with a very high-level overview of the primary modules and progressively provide details on where our contributions (the novel reward function, the cover detection algorithm, the integrated cover following navigation planning) fit in.
% and progressively dive deeper into the intricacies of these module in the following subsections. 
% Our hypothesis is that maintaining a close distance to nearby obstacles like trees while avoiding collisions is the gist to safely find maximum cover in unstructured outdoor environments. We also think that avoiding elevation gain will help the robot stay unnoticed.
% However, if no shelters are present near the robot, or if nearby shelters are in unsafe regions, the robot should navigate directly to the predefined location by planning a short and safe trajectory. Then, it follows the path with dynamically feasible velocities. Our approach adapts and integrates the following existing algorithms for the perception of the environment and the navigation planning model.

\subsection{Overview of \textit{CoverNav} Navigation Planning}
%-----------------------------------------------------------------
The \textit{CoverNav} navigation planning system (Figure~\ref{coverNav}) comprises three interdependent modules that enable a mobile robot to navigate safely through unstructured outdoor environments in search of cover while avoiding obstacles and remaining undetected. The details of the modules are articulated below:

% \begin{enumerate}[leftmargin=*]
% \item \textbf{Perception and Semantic Segmentation:} 
\subsubsection{Perception and Semantic Segmentation}
This module receives and processes image and point cloud sensor data from the environment. We first infer the robot's pose, twist, and elevation map from this data. Semantic segmentation techniques are applied to raw sensor data to detect and classify objects in the environment.
% As presented in Listing~\ref{lst:detection},

% our algorithm determines which objects should serve as covers to enable the robot to navigate safely to a predefined goal location. This process takes into account the distance of objects from the robot, as well as the potential for collisions and the amount of cover provided.

% \item \textbf{Cover Analysis and Selection:}
\subsubsection{Cover Analysis and Selection}
As the semantic segmentation data is passed to this module, we utilize our novel cover detection Algorithm (described later Section~\ref{sec:sementic_seg_cover-detect}) to identify potential cover providing natural obstacles such as trees, rocks, bushes, etc.
This process takes into account the distance of objects from the robot, as well as the potential for collisions and the amount of cover provided.
Once potential cover objects are identified, this module calculates the 3D Euclidean distance between the robot and each potential cover object. It then applies predefined selection criteria, including factors such as distance, and visibility, to determine the most suitable cover object. 
% The objective is to optimize the reward function, encouraging the robot to seek cover objects that offer maximum concealment and minimize visibility.

% \item \textbf{Path Planning:} This module uses the TERP+Cover Following algorithm to generate a trajectory that guides the robot towards the selected cover object, while avoiding obstacles. The computed trajectory takes into consideration the robot's current position, the cover object's location, and the surrounding environment. The goal is to determine a safe and optimal trajectory that maximizes cover, minimizes visibility, and assures obstacle avoidance. This trajectory is dynamically updated in response to changes in the environment or the detection of an adversary.

% \item \textbf{Motion Control:} The final module harnesses the generated trajectory to control the robot's actuators (e.g., motors or wheels) to execute the planned movements. It employs the DWA-RL algorithm to determine dynamically feasible velocities that enable the robot to adhere to the trajectory while avoiding collisions and maintaining stability. The DWA-RL algorithm merges the strengths of the Dynamic Window Approach (DWA) and Deep Reinforcement Learning (DRL) methods to facilitate safe and efficient navigation in dynamic environments with mobile obstacles.

\subsubsection{Path Planning and Motion Control}
% \item \textbf{Path Planning and Motion Control:} 

% a Deep Reinforcement Learning (DRL) network based on the Deep Deterministic Policy Gradient (DDPG)~\cite{lillicrap2015continuous} algorithm. The DRL network is composed of Convolutional Neural Network (CNN) layers that enable it to learn the relationship between the robot's inputs and the optimal cover-following velocities required for safe navigation.

This module integrates the path planning and motion control functionalities to facilitate effective navigation of the robot. In this module, we combine TERP~\cite{weerakoon2022terp} (a Deep Deterministic Policy Gradient (DDPG)~\cite{lillicrap2015continuous} algorithm) with our novel reward function to generate a trajectory that guides the robot towards the selected cover object, while avoiding obstacles.
The computed trajectory takes into consideration the robot's current position, the cover object's location, and the surrounding environment. The goal is to determine a safe and optimal trajectory that maximizes cover, minimizes visibility, and assures obstacle avoidance. This trajectory is dynamically updated in response to changes in the environment or the detection people within the environment. The motion control aspect utilizes the DWA-RL~\cite{patel2020dynamically} algorithm, which combines the strengths of the Dynamic Window Approach (DWA)~\cite{fox1997dynamic, seder2007dynamic} and Deep Reinforcement Learning (DRL) methods. It determines dynamically feasible velocities that enable the robot to adhere to the trajectory while ensuring collision avoidance and maintaining stability.\\
% By integrating path planning and motion control, the robot can navigate efficiently and safely in dynamic environments with mobile obstacles, maximizing cover and minimizing visibility.
% \end{enumerate}

\begin{figure}[htbp]
  \centering
  \includegraphics[width=0.5\textwidth]{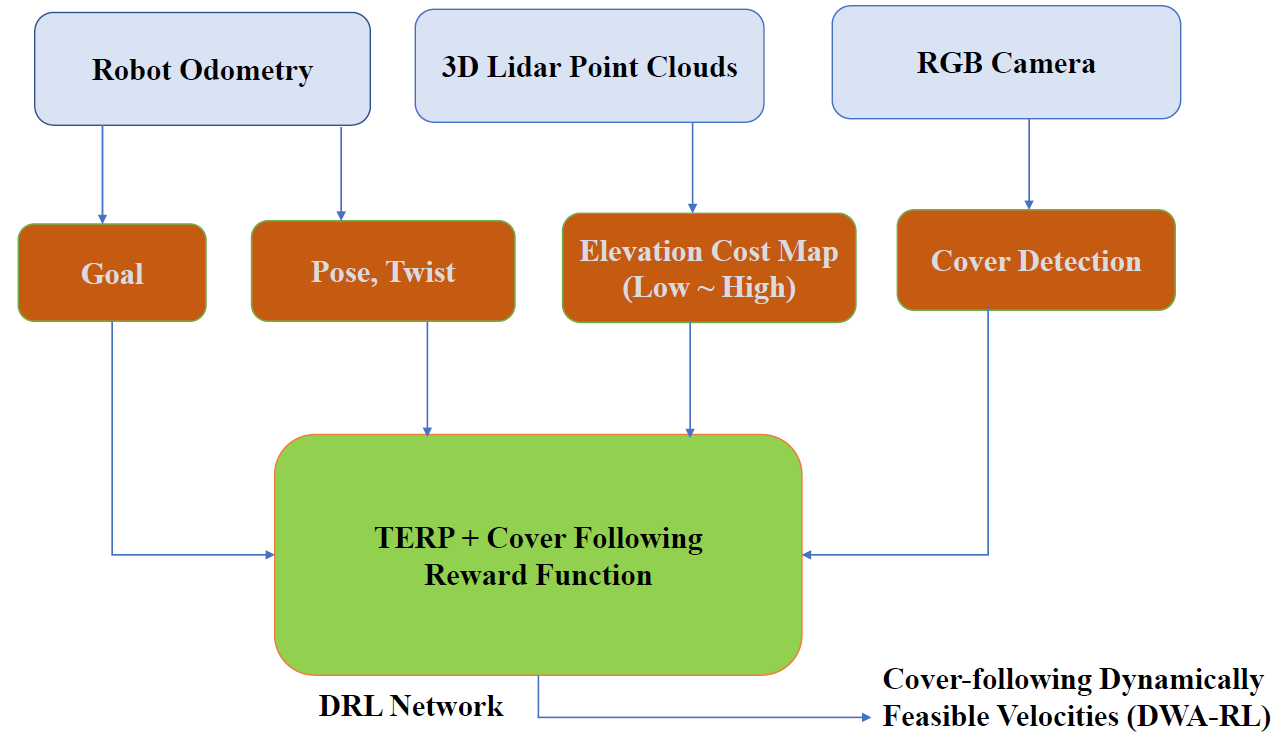}
  \caption{ Overview of \textit{CoverNav} System Architecture.}
  \label{sysarch} 
\end{figure}

These three modules work together to enable the robot to autonomously and intelligently navigate towards its goal, maximizing safety, and minimizing the risk of detection.
The system architecture is shown in Fig.~\ref{sysarch}, where the inputs from the sensors are processed through different modules to produce the output of dynamically feasible velocities for the robot to follow. The cover detection and reward function modules use semantic segmentation information to identify potential covers for the robot to use, and the reward function provides feedback to the DRL network for optimizing the cover-following velocities. The elevation cost map provides information about the terrain and helps the robot avoid elevation gains that may make it more visible.

% \begin{figure}[htbp]
% \centerline{
% \includegraphics[width=0.5\textwidth]{images/system_architecture_3.png}}
% \caption{ Overview of \textit{CoverNav} System Architecture.}
% \label{sysarch}
% % \vspace{-2ex}
% \end{figure}

% The output of the DRL network is then fed into the motion planning module, which uses the TERP+Cover Following algorithm to generate a safe and optimal trajectory for the robot to follow towards the goal location. The TERP+Cover Following algorithm takes into account the robot's desired velocity, the surrounding environment, and the amount of cover available, to create a smooth and collision-free trajectory.

\subsection{Problem Formulation}
Navigating a UGV in unstructured outdoor environments with covers poses a challenging problem due to the need for obstacle avoidance, cover detection, and reliable path planning. The objective is for the robot to approach a designated goal location, while ensuring obstacle avoidance, pose stability, and maintaining proximity to available covers. In order to effectively tackle the aforementioned challenge, we must account for various scenarios and uncertainties. Therefore, we formulate our problem as a Markov decision process (MDP). The MDP formulation allows us to model the environment, the robot's actions, and the resulting outcomes in a probabilistic manner, enabling us to develop an optimal policy that maximizes the expected reward over time. Our MDP consisting of a tuple $(S, A, P, R, \gamma)$, where:\\

\vspace{-1ex}
\begin{itemize}[leftmargin=*]
    \item $S$ is the state space representing the robot's state. It includes the robot's position and orientation, as well as the presence and location of obstacles and covers in the environment.
    \item $A$ is the action space representing the robot's actions. It includes the robot's movement in different directions and its rotation.
    \item $P$ is the transition function representing the probability of moving from one state to another after taking an action.
    \item $R$ is the reward function representing the immediate reward the robot receives after taking an action.
    \item $\gamma$ is the discount factor representing the importance of future rewards.
\end{itemize}
Our objective is to find a policy $\pi(s)$ that maps a state $s$ to action $a$, such that the expected cumulative reward of the robot is maximized. 

% \noindent \textbf{Reward Function Design:} 
\subsubsection{Reward Function Design}
Our reward function is designed to encourage desirable robot behaviors and discourage undesirable ones. It includes five components:

% \begin{enumerate}[leftmargin=*]
%    \item \textbf{$r_{goal}$:}
% \end{enumerate}

\begin{description}[leftmargin=1em]
    \item[$r_{goal}$:]  This term measures progress towards the goal by subtracting the current distance from the previous distance. The closer the robot gets to the goal, the higher the reward. It is calculated as the difference between the distance to the goal in the previous state ($d_{t-1}$) and the distance to the goal in the current state ($d_t$). A positive reward is assigned if the robot moves closer to the goal, indicating progress, while a negative reward is assigned if the robot moves farther away.
    \begin{equation}
       r_{\text{goal}} = d_{t-1} - d_t
    \end{equation}
    \item[$r_{dir}$:] This reward component encourages the robot to maintain a consistent heading toward the goal. It is calculated as the absolute difference between the heading in the current state ($\theta_t$) and the heading in the previous state ($\theta_{t-1}$). A smaller difference indicates a more consistent heading, resulting in a higher reward.
    \begin{equation}
        r_{\text{dir}} = -|\theta_t - \theta_{t-1}|
    \end{equation}
    \item[$r_{stab}$:]  This reward component evaluates the stability of the robot based on its roll and pitch angles ($\psi_t$). The reward is calculated using an exponential function, where the stability reward decreases as the roll and pitch angles deviate from zero. Smaller values of $\psi_t$ result in higher stability rewards.

    \begin{equation}
         r_{\text{stab}} = \exp(-\psi_t^2)
    \end{equation}
    \item[$r_{elev}$:] This term encourages the robot to move towards areas of higher elevation by penalizing deviations from the target elevation. It is computed as the sum of absolute differences between the current and target elevations, weighted by a vector of elevation weights. Mathematically, it can be represented as:

    \begin{equation}
    r_{elev} = \sum_{i=1}^{n} w_{elev}\Delta h_i
    \end{equation}
    
    \noindent where $n$ is the number of previous positions, $\Delta h_i$ is the difference in elevation between the current position and the $i^{th}$ previous position, and $w_{elev}$ is the weight given to the elevation component.    
    The calculation of $\Delta h_i$ depends on the method used to obtain elevation information. In this work, the elevation information is obtained from the Unity terrain data. $\Delta h_i$ is calculated as the difference between the elevation at the current position and the $i^{th}$ previous position, given by:
    
    \begin{equation}
    \Delta h_i = h_{cur} - h_{i}
    \end{equation}
    
    where $h_{cur}$ is the elevation at the current position and $h_i$ is the elevation at the $i^{th}$ previous position.

    \item[$r_{cover}$:] This term encourages the robot to seek cover objects to hide behind or beneath. It is computed as follows:
    
    \begin{equation}
    \footnotesize
        \begin{aligned}
        r_{\text{cover}} =
        \begin{cases}
        0, & \text{if } d_{\text{coverObject}} > 1.5w_{\text{min}} \\
        d_{\text{coverObject}} - 0.5w_{\text{min}}, & \text{if } 0.5w_{\text{min}} \leq d_{\text{coverObject}} \leq 1.5w_{\text{min}} \\
        -1000, & \text{if } d_{\text{coverObject}} < 0.5w_{\text{min}}
        \end{cases}
        \end{aligned}
    \end{equation}
     It is calculated based on the distance to the nearest cover object, represented by $d_{\text{coverObject}}$, and the shortest external dimension of the robot, denoted as $w_{\text{min}}$. If the distance to the nearest cover object ($d_{\text{coverObject}}$) is greater than 1.5 times the shortest external dimension of the robot, then the reward is 0.
    If the distance to the nearest cover object is between 0.5 and 1.5 times the shortest external dimension of the robot, then the reward is the distance minus half the shortest external dimension of the robot.
    If the distance to the nearest cover object is less than 0.5 times the shortest external dimension of the robot, then the reward is -1000.
    The reward function is designed to encourage the robot to move towards the goal while avoiding obstacles and seeking cover. The different terms provide incentives for different behaviors, such as moving in a straight line, maintaining stability, and seeking higher ground. The weighting of the terms can be adjusted to optimize the performance of the robot.
\end{description}
The reward at current state $R_t$ is the sum of these five components, summarized at the end of Table~\ref{tab:reward-function}.

\begin{table}[!htb]
    \centering
    \begin{tabular}{|lcl|}
    \hline
    $d_t$ & & distance from robot to goal at current state \\\hline
    $d_{t-1}$ & & distance from robot to goal at previous state \\\hline
    $\theta_t$ & & heading of the robot at current state \\\hline
    $\theta_{t-1}$ & & heading of the robot at previous state \\\hline
    $\psi_t$ & & roll and pitch angles of the robot at current state \\\hline
    $w$ & & weight vector for elevation changes \\\hline
    $d_{\text{coverObject}}$ & & distance to the nearest cover object \label{line:ct} \\\hline
    $w_{min}$ & & shortest external dimension of the robot \\\hline
    $r_{goal}$ &=& $d_{t-1} - d_t$ \\\hline
    $r_{dir}$ &=& $-|\theta_t - \theta_{t-1}|$ \\\hline
    $r_{stab}$ &=& $\exp(-\psi_t^2)$ \\\hline
    $r_{elev}$ &=& $\sum_i w_elev |\Delta h_i|$ \\\hline
    $r_{cover}$ &=& $\tiny\begin{cases}
        0, & \text{if } d_{\text{coverObject}} > 1.5w_{\text{min}} \\
        d_{\text{coverObject}} - 0.5w_{\text{min}}, & \text{if } 0.5w_{\text{min}} \leq d_{\text{coverObject}} \leq 1.5w_{\text{min}} \\
        -1000, & \text{if } d_{\text{coverObject}} < 0.5w_{\text{min}}
        \end{cases}$\\\hline
    $\bm{R_t}$ &=& $\bm{r_{goal} + r_{dir} + r_{stab} + r_{elev} + r_{cover}}$ \\\hline
    \end{tabular}
    \vspace{2pt}
    \caption{Summary of Reward Function Components}
    \label{tab:reward-function}
\end{table}

% \begin{algorithm}
% \caption{Reward function}
% \label{alg:reward-function}
% \begin{algorithmic}[1]
% \REQUIRE \Statex $s$: current state of the robot
% \REQUIRE \Statex $s'$: next state of the robot
% \REQUIRE \Statex $\mathcal{C}$: set of nearby covers
% \REQUIRE \Statex $\mathcal{O}$: set of obstacles
% \REQUIRE \Statex $\beta_i$: weight factor for the $i$-th component of the reward function \\
% \STATE $R_{\mathrm{distance\ rate}} \leftarrow d_{\mathrm{past\ dist}} - d_{\mathrm{current\ dist}}$ \\
% \STATE $R_{\mathrm{heading}} \leftarrow -|\alpha_{\mathrm{goal}}|$ \\
% \STATE $R_{\mathrm{elevation}} \leftarrow -w \cdot h$ \\
% \STATE $R_{\mathrm{cover}} \leftarrow d_{\mathrm{coverObject}}$ \\
% \STATE $R_{\mathrm{stability}} \leftarrow \cos^2\phi + \cos^2\theta$ \\
% \STATE $R_{\mathrm{total}} \leftarrow \beta_1 R_{\mathrm{distance\ rate}} + \beta_2 R_{\mathrm{heading}} + \beta_3 R_{\mathrm{elevation}} + \beta_4 R_{\mathrm{cover}} + \beta_5 R_{\mathrm{stability}}$
% \RETURN $R_{\mathrm{total}}$
% \end{algorithmic}
% \end{algorithm}

\subsubsection{Cover Detection}
\label{sec:sementic_seg_cover-detect}
To accurately locate and label objects, we utilize the ARL Unity Framework's semantic segmentation framework. This algorithm analyzes the scene and assigns relevant class labels to each pixel in the image. The resulting segmentation map provides valuable information about the different objects present, including trees, buildings, rocks, and grass (refer to Listing~\ref{lst:detection} for an example).
However, to specifically identify objects that can serve as cover for the robot, we apply additional criteria to the semantic labels. We focus on objects that have the potential to offer protection and concealment, such as buildings, trees, junk cars, etc. By narrowing down the selection to these cover objects, we ensure that our system targets the most suitable options for the robot's safety and strategic positioning.
Once we have identified the desired cover object, we proceed to calculate the distance between the robot's position and the cover object in a 3-dimensional space. This distance metric provides crucial information about the proximity of the cover and helps the robot in making informed decisions during navigation and planning.

To calculate the distance, we use the 3D Euclidean distance formula. In the context of the reward function, this formula is used to calculate the distance between the robot and a cover object in 3D space, as shown in Equation ~\eqref{eq:distance}:
% \begin{multline*}
% \label{eq:distance}
% d_{coverObject} = \sqrt{(x_{robot} - x_{cover})^2 + (y_{robot} - y_{cover})^2 \\ + (z_{robot} - z_{cover})^2}
% \end{multline*}
\begin{equation}
\label{eq:distance}
\begin{aligned}
d_{\text{coverObject}}
    &= \sqrt\big(
    (x_{\text{robot}} - x_{\text{cover}})^2 \\
    &+ (y_{\text{robot}} - y_{\text{cover}})^2 + (z_{\text{robot}} - z_{\text{cover}})^2
    \big)
\end{aligned}
\end{equation}
% \begin{equation}
% \label{eq:distance}
% \begin{split}
% d_{\text{coverObject}} = \sqrt{&(x_{\text{robot}} - x_{\text{cover}})^2} \\
% &+ (y_{\text{robot}} - y_{\text{cover}})^2 + (z_{\text{robot}} - z_{\text{cover}})^2
% \end{split}
% \end{equation}
where $d_{\text{coverObject}}$ represents the distance between the robot and the cover object, and $(x_{\text{robot}}, y_{\text{robot}}, z_{\text{robot}})$ and $(x_{\text{cover}}, y_{\text{cover}}, z_{\text{cover}})$ represent the 3D coordinates of the robot and the cover object, respectively. If the system is able to recognize any of the relevant cover categories, such as trees, bushes, or rocks, they are passed on to Algorithm~\ref{CoverAlgo}. 

% If we are able to recognize any of the relevant cover categories, we  send them on to Algorithm~\ref{CoverAlgo} Line~\ref{line:ct} (for training, which includes the process of rewarding participants.

% \begin{figure}[htbp]
% \centerline{
% \includegraphics[width=0.45\textwidth]{images/ss_info.png}}
% \caption{Semantic Segmentation Object Information}
% \label{objectInfo}
% \end{figure}

% frame_id: 991904 
% detections:
%     class_id: 0
%     class_name: "Rock"
%     confidence: 1.0
%     x_min:151.64044189453125
%     y_min:118.20899963378906
%     x_max: 168.4332275390625
%     y_max: 112.97318267822266 
%     object_id: 68
%     x_pos: -0.18053817749023438 
%     y_pos: -0.46726706624031067 
%     z_pos: 18.7108097076416 
%     kpts: []

\begin{lstlisting}[caption={Detected Object Information},captionpos=b,label={lst:detection},style=yaml]
frame_id: 991904 
detections:
    class_id: 0
    class_name: ''Rock''
    confidence: 1.0
    x_min: 151.64044189453125
    y_min: 118.20899963378906
    x_max: 168.4332275390625
    y_max: 112.97318267822266
    object_id: 68
    x_pos: -0.18053817749023438
    y_pos: -0.46726706624031067
    z_pos: 18.7108097076416
    kpts: []
\end{lstlisting}

% \begin{minted}[
%     % gobble=4,
%     frame=single,
%     % linenos
%   ]{yaml}
% frame_id: 991904 
% detections:
%     class_id: 0
%     class_name: "Rock"
%     confidence: 1.0
%     x_min: 151.64044189453125
%     y_min: 118.20899963378906
%     x_max: 168.4332275390625
%     y_max: 112.97318267822266
%     object_id: 68
%     x_pos: -0.18053817749023438
%     y_pos: -0.46726706624031067
%     z_pos: 18.7108097076416
%     kpts: []
% \end{minted}

Given the details of detected objects, such as object class, object location, robot location, and object detection confidence, Algorithm~\ref{CoverAlgo}  determines whether the robot is currently in cover or not. Algorithm~\ref{CoverAlgo} operates by iterating through each detected object and computing the distance between the object and the robot. The algorithm considers objects that belong to classes such as trees, bushes, rocks, etc., and has a detection confidence score of at least 0.85. The confidence threshold of 0.85 ensures that only objects with a sufficiently high detection confidence are considered for cover detection. The algorithm stores the minimum distance among these objects in the variable \texttt{coverDistance}. If the minimum distance is less than or equal to 10 meters, the algorithm returns True, indicating that the robot is currently in cover. Otherwise, the algorithm returns False, indicating that the robot is not in cover. The cover detection algorithm offers a time complexity of O(N), where N is the number of detected objects. The algorithm efficiently iterates through the detected objects, computes distances, and determines if the robot is in cover based on specified thresholds. The chosen threshold values, such as detection confidence and distance, influence the object selection and cover determination process without affecting the algorithm's time complexity. 

\begin{algorithm2e}%[H]
\SetAlgoLined
\DontPrintSemicolon
% \KwIn{$objectClass,objectLocation, robotLocation,confidence$\Comment{Details of detected objects}}
\KwIn{$objC$\Comment{Object Classes}}
\KwIn{$objL$\Comment{Object Locations}}
\KwIn{$rL$\Comment{Robot Location}}
\KwIn{$cf$\Comment{Confidence Values}}
\KwOut{$isCover$ \Comment{Cover or not cover}}
\SetKwFunction{FMain}{DetectCover}
\SetKwProg{Fn}{Function}{:}{}
\Fn{\FMain{$objC,objL, rL, cf$}}{
    $ coverDistance  \longleftarrow \infty $;\\
    $isCover \longleftarrow False$\\
    $N \longleftarrow length(objC)$\\
    \ForEach{$ i \in N $}{
        \If{
            $objC_i \in \{trees, bushes, rocks, cottage, building, houses,\newline disabled~vehicles\}$~\textbf{and} $ cf_i \geq 0.85 $}
        {
            $ dist_i \longleftarrow ||objL_i - rL|| $\\
            \If{$ dist_i < coverDistance $}{
                $ coverDistance \longleftarrow dist_i $\\
            }
        }
    }
    \If{$ coverDistance \leq 10 $}{
        $ isCover \longleftarrow True $\\
    }
    \textbf{return} $isCover; $ 
}
% \textbf{End Function}
\caption{Algorithm for Detecting Cover}
\label{CoverAlgo}
\end{algorithm2e}

\subsubsection{Dynamic Window Approach with Deep Reinforcement Learning}

The Dynamic Window Approach (DWA)~\cite{fox1997dynamic, seder2007dynamic} is a well-known navigation algorithm that takes into account the dynamic constraints of a robot and ensures collision-free and feasible velocities within a given time interval. It operates within a 2-dimensional velocity space known as the dynamic window, which contains a discrete set of velocity vectors that the robot can achieve in the next time interval. The DWA algorithm consists of two stages to determine a suitable velocity vector. Firstly, it evaluates the collision-free velocities within the dynamic window by considering the robot's dynamic constraints. This stage ensures that the selected velocities are within the robot's capability to execute safely. However, the DWA algorithm has limitations when it comes to navigating in the presence of mobile obstacles. It relies solely on the robot's immediate sensor data, which may not provide enough time for the robot to react to changes in the environment. As a result, it may struggle to effectively avoid collisions with moving obstacles.

To overcome this problem, we employ a recent work~\cite{patel2020dynamically} that combines the advantages of DWA and DRL-based methods for navigation in dynamic environments where mobile obstacles exist. This method functions as our navigation module, which is advantageous to our method in a number of different respects. First, it contains a reward function that is shaped such that the robot’s navigation is more spatially aware of the obstacles’ motion. That is, the robot is rewarded for navigating in the direction opposite to the heading direction of obstacles, which leads to the robot taking maneuvers around moving obstacles. Second, using our DRL agent in conjunction with a high-level path planner like this can reduce the complexity of the DRL problem and speed up the training process. That means, calculating the robot's feasible velocity set at a particular time instant and the costs that correspond to making use of that robot's velocity over the course of the previous n time instants is how the observation space is built up in DWA-RL~\cite{patel2020dynamically}. This formulation maintains the dynamic feasibility guarantees provided by DWA while also incorporating the time-dependent change in the state of the environment. In contrast to the outcomes of other DRL methods, this one results in a significantly reduced dimension of the observation space.

\subsection{Training Strategy}

The training strategy we employ is a combination of exploration with a DRL agent and exploitation of DWA's dynamic feasibility guarantees. In the beginning, the DRL agent explores the environment, using its policy to determine the robot's action. Exploration is done using random actions to enable the agent to learn the optimal policy. Once a sufficient number of episodes have been run, the DRL agent is tested with DWA. During testing, DWA's dynamic feasibility guarantees ensure that the robot does not collide with obstacles while moving towards the goal.

The learning rate for both actor and critic networks was set to $10^{-4}$. The batch size was set to 64, and the replay buffer size was set to 100,000. We use the Adam optimizer with a decay rate of 0.9 and 0.999 for the first and second moments, respectively. The discount factor $\gamma$ was set to 0.99, and the exploration noise scale $\sigma$ was set to 0.1.

The training process consists of a total of 100 episodes. Each episode lasts for 100 time-steps, during which the robot navigates through the environment, avoiding obstacles and maximizing cover while minimizing visibility. At the beginning of each episode, the robot's initial position is randomized within the start zone. During each time step, the robot receives an observation from the environment, and the DRL agent selects an action based on the observation. The action is then executed in the environment, and the robot's new state is observed, along with the reward. The process repeats until the episode ends, either because the robot has reached the goal or has exceeded the time limit. We made the decision to stop the training at episode 100 after careful analysis and consideration of various factors. By episode 100, the agent had already shown significant progress and achieved competitive performance. Further training beyond this point did not yield substantial improvements in terms of success rate and trajectory length. Considering the computational resources required for continued training, we determined episode 100 as an appropriate stopping point.

The reward function is designed to encourage the robot to move towards the goal while maximizing cover and minimizing visibility. The maximum cover and minimum visibility that can be obtained during the episode are tracked, and the final reward is normalized by these values. This normalization encourages the robot to achieve maximum cover and minimum visibility during each episode.

\section{Experiments and Results}
\label{sec:experiment}
% \begin{figure}[htbp]
% \centerline{
% \includegraphics[width=0.45\textwidth]{images/arl.png}}
% \caption{The ARL Unity Simulation environment}
% \label{fig}
% \end{figure}
We explain the implementation of our technique as well as the experimental details on a Husky robot in the ARL unity simulation environment. 
\subsection{Implementation Details}

The model is tested in the ARL Unity Simulation~\cite{fink2019arl}, a realistic outdoor environment that has a broad range of ground textures and elevation levels, as well as buildings, vegetation, lake, etc. Our DRL network is implemented in PyTorch. We use simulated uneven terrains with a Clearpath Husky robot
created using ROS Melodic~\cite{quigley2009ros} and ARL Unity Simulation framework to train the DRL network. The simulated Husky robot is mounted with a Velodyne VLP16 3D LiDAR. The network is trained in a workstation with a 10th Generation Intel Core i9-10850K processor and an NVIDIA GeForce RTX 3090 GPU.

We utilize the existing semantic segmentation framework from ARL Unity simulation and apply some custom labeling to it. The custom labeling includes ignoring small objects that cannot provide cover for the robot, which has been done to serve the specific purposes of our experiment. We use the Octomap~\cite{hornung2013octomap} and Grid-map~\cite{fankhauser2016universal} ROS packages to obtain the elevation map using the point clouds from the Velodyne ROS package. The computational load of our network is low enough that it can function effectively in real-time on the laptop with the ROS tools that were discussed earlier.

\subsection{Experiment Details} 

The following details provide insights into our experimental setup.
 
\textbf{Number of Training Episodes:} We conduct extensive training sessions consisting of 100 episodes to ensure thorough exploration and learning. Each episode represents a complete navigation mission, allowing the robot to learn and adapt to various environmental conditions. We empirically found that 100 is an adequate number of episodes to ensure robust learning and the acquisition of diverse navigation strategies.

\textbf{Destination Placement:} The destinations for the navigation missions are strategically set within a radius of 12 meters from the robot's initial location. This range provides a challenging yet realistic scenario for the robot to navigate to different targets. It enables us to assess the robot's capability to explore and reach destinations in various environmental conditions, including terrain variations and obstacle configurations.

\textbf{Reward Scores:} The reward scores obtained during the training sessions serve as a metric to evaluate the performance improvement of our model over time. We monitor the average reward score throughout the training process. The gradual increase in reward scores indicates the learning progress and the effectiveness of our method in achieving navigation goals. Higher reward scores demonstrate improved navigation capabilities and successful adherence to cover-following strategies.

\subsection{Baselines}
\label{baselines}
We compare the performance of our algorithm against the following navigation strategies:
\begin{itemize}[leftmargin=*]
    \item \textbf{DWA~\cite{fox1997dynamic}:} A popular algorithm that focuses on real-time obstacle avoidance and trajectory planning. It calculates feasible velocities based on the current robot state and dynamically selects the best velocity combination to avoid collisions and reach the goal. 
    \item \textbf{GA-Nav~\cite{guan2022ga}:} A learning-based semantic segmentation method that aims to identify different terrain groups in unstructured and outdoor environments. It has shown promising results in accurately classifying terrain surfaces and improving navigation performance in complex terrains.
    \item \textbf{TERP~\cite{weerakoon2022terp}:} A DRL-based planner specifically designed for navigating uneven outdoor terrains. It incorporates terrain elevation information to identify unsafe regions and computes local least-cost waypoints to avoid traversing those areas.
\end{itemize}
% (i) (ii)  (iii) 
% \textbf{INTRODUCE THE BASELINES FIRST, THEN DESCRIBE IN DETAIL WHAT YOU SEE IN THESE TWO FIGURES.}

\subsection{Evaluation Metrics}
\label{evaluation_metrics}
% The evaluation of our method's navigation performance is based on several key parameters, providing valuable insights into the learning process and the effectiveness of our approach.

% By considering these evaluation parameters, we gain valuable insights into the learning curve, the adaptability of our approach to different scenarios, and the overall navigation performance of our method. These evaluations provide a comprehensive assessment of our method's effectiveness and robustness in real-world navigation scenarios. 

Due to fundamental differences between some of the baselines and \textit{CoverNav} (e.g., neither DWA and GA-Nav are DRL-based methods), direct comparison of reward scores between \textit{CoverNav} and the baseline methods is not meaningful. Instead, we focus on evaluating navigation performance using the following metrics for a fair comparison. 
% We use the following metrics to evaluate our method’s navigation performance.

% The short training sessions contain around 50 episodes, whereas an episode means all states from the initial state to the final state within one navigation mission, reflect fast and steady learning curves. The destinations are also set to be in a radius of 12 meters from the robot’s initial location.

% The reward value in these training sessions gradually goes up as expected and it is depicted in (Fig. \ref{learningCurve}).

% The initial reward score as well as the average reward score demonstrates significant signs of improvement in performance as the number of episodes played increases as can be seen in (Fig. \ref{orgvsavg}). Additionally, we use the following metrics to evaluate our method’s navigation performance.
% \begin{figure}[htbp]
% \centerline{
% \includegraphics[width=0.5\textwidth]{images/reward_vs_average_reward_100.png}}
% \caption{Original reward vs average reward score in each episode}
% \label{orgvsavg}
% \end{figure}

% \begin{figure}[htbp]
% \centerline{
% \includegraphics[width=0.45\textwidth]{images/success_rate_fig.png}}
% \caption{Success rate of achieving goal in various scenarios}
% \label{successRate}
% \end{figure}

\textbf{Success Rate:} The ratio (in percentage) between the number of times the robot successfully accomplished its objective and the number of total tries, while avoiding obstacles, high elevations, going through cover, and low elevations. 

% The percentage of successful outcomes was determined based on five, ten, and twenty tests. Fig. \ref{successRate} depicts the percentage of times a goal is accomplished successfully across a variety of contexts.

% \begin{figure*}[h!]
%     \centering
%     \subfloat[\centering ]{{\includegraphics[width=0.35\textwidth]{images/robot_training_01.JPG}}}%
%     \hfill
%     \subfloat[\centering ]{{\includegraphics[width=0.35\textwidth]{images/robot_training_02.JPG}}}%
%     \hfill
%     \subfloat[\centering]{{\includegraphics[width=0.35\textwidth]{images/robot_training_03.JPG}}}%
%     \hfill
%     \subfloat[\centering ]{{\includegraphics[width=0.35\textwidth]{images/robot_training_04.JPG}}}%
%     \caption{A step-by-step look at how the robot is finding the shelter. (a) the robot encountered an obstacle, (b) heading towards the goal avoiding the obstacle, (c) finding a position near to the goal, and (d) reached the goal location.}%
%     \label{huskyTraining}
% \end{figure*}

\begin{figure*}[h!]
    \centering
    \subfloat[\centering ]{{\includegraphics[width=8cm]{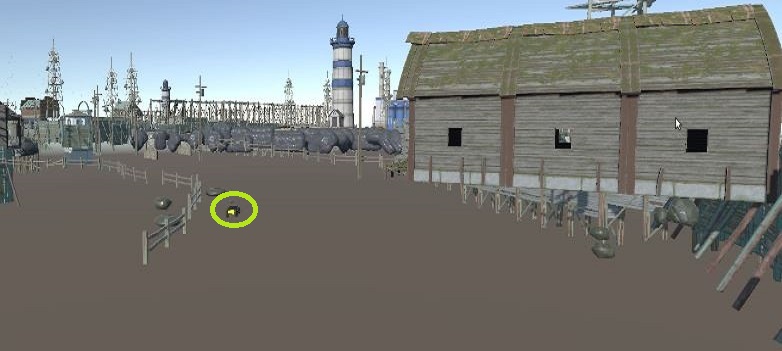} }}%
    \qquad
    \subfloat[\centering ]{{\includegraphics[width=8cm]{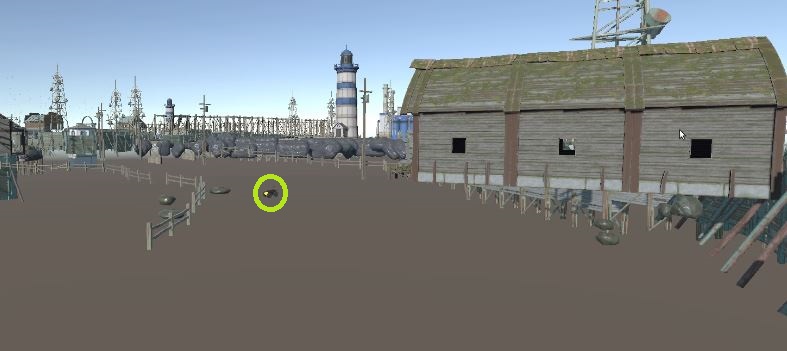} }}%
    \qquad
    \subfloat[\centering]{{\includegraphics[width=8cm]{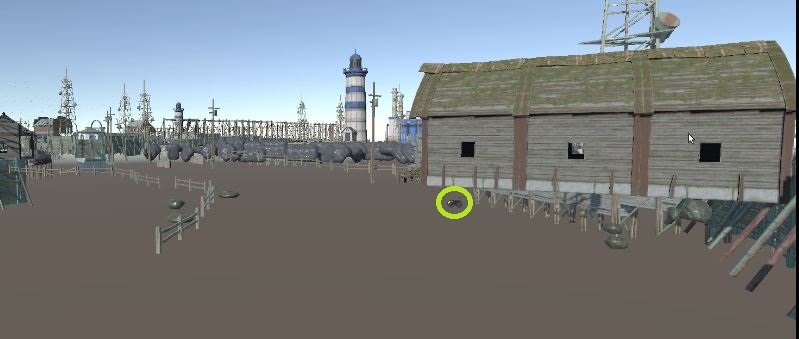} }}%
    \qquad
    \subfloat[\centering ]{{\includegraphics[width=8cm]{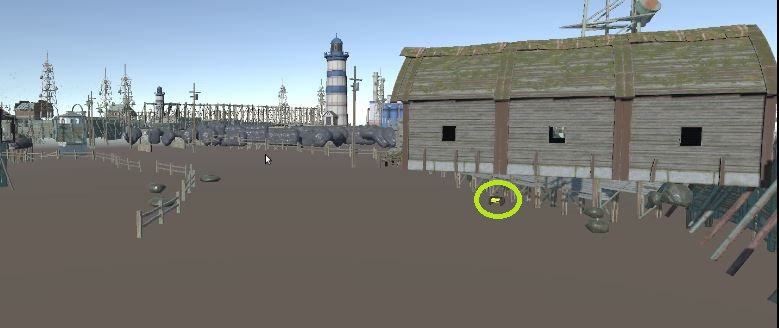} }}%
    \caption{A sequential overview of the robot's process in finding shelter. (a) The robot initially encounters an obstacle along its intended path. (b) It navigates around the obstacle while still heading towards the goal location. (c) The robot identifies and selects a strategic position near the goal that offers sufficient cover or shelter. (d) Finally, the robot successfully reaches the goal location, utilizing the nearby cottage as a shelter or cover during its navigation. The step-by-step visualization demonstrates the robot's adaptive behavior in identifying and utilizing available shelter to accomplish its objectives.}%
    \label{huskyTraining}
\end{figure*}
% \textbf{Collision Rate}- The proportion of episodes in which the robot runs into an obstruction to all of the trials combined.

% \textbf{Timeout Rate}- The ratio of the number of times, excluding collision scenarios, that the robot fails to complete the task within the allotted steps.

\textbf{Execution time:} The time (in seconds) taken by the robot to complete a navigation task. Shorter execution times indicate a more efficient and responsive navigation system, making it suitable for time-sensitive applications. Optimizing execution time is crucial for practical and effective navigation in real-world scenarios.

\textbf{Trajectory length:} The distance (in meters) traveled by the robot to reach its destination. This metric provides insights into the efficiency of navigation. Shorter trajectories indicate more efficient navigation, while longer trajectories may suggest suboptimal paths or difficulty navigating in a complex environment.

% The results show that our proposed DRL-based navigation system achieves a high success rate in various scenarios. For instance, in the low-high elevation environment, the success rate is 80\% for five tests, 60\% for ten tests, and 50\% for twenty tests. In the forest and jungle environment, the success rate is 70\% for five tests, 50\% for ten tests, and 40\% for twenty tests. Fig. \ref{successRate} shows the success rates achieved for each scenario.

% \subsubsection{Add a header!}
% It can be observed that the agent consistently improves its performance across all scenarios as the number of episodes progresses, demonstrating effective learning

%\subsubsection{Add a Title}
% \begin{figure}[!htb]
% \centerline{
% \includegraphics[width=0.5\textwidth]{images/reward_vs_average_different_environment.png}}
%     \caption{The Reward vs. Average Reward for different scenarios over a series of episodes. Each subplot corresponds to a different scenario: 'GR (Low)', 'OA (Normal)', 'UT (Low-High)', 'CF'.}
% \label{orgvsavg}
% \end{figure}

% \subsection{Comparison with State-of-the Methods?}
% We evaluate our cover-following based navigation in different environment scenarios. Fig. \ref{huskyTraining} shows in-depth look at the methodical process by which the robot is attempting to locate a place to stay while using the house as shelter.

\subsection{Testing Scenarios}
\label{testing_scenario}
We evaluate our \textit{CoverNav} navigation in different environment scenarios, including normal elevation, low-high elevation, and forest and jungle environments.
\begin{itemize}[leftmargin=*]
    \item \textbf{Normal Elevation:} This refers to a terrain with a relatively flat or gently sloping surface.
    \item \textbf{Low Elevation:} The terrain has a normal height and features such as buildings, fences, and other minor impediments. The maximum elevation ($\leq$ 1m). 
    \item \textbf{Low-High Elevation:} The terrain includes both high and low elevations. It has elevations ranging from 1m to 3m.
    \item \textbf{Forest and Jungle:} A realistic forest environment that includes diverse terrain and extensive areas of cover. The diverse terrain includes grass, an open hiking path, water, and some hills\cite{ForestSimulation2022}. The maximum elevation $\geq$ 4m. 
\end{itemize}

\subsection{Results}

In this section, we analyze the navigation performance of \textit{CoverNav} in various testing scenarios mentioned in the previous section.
% Section~\ref{testing_scenario}
We assess \textit{CoverNav}'s ability to navigate through different terrains, overcome obstacles, and locate suitable shelter spots. Finally, we compare the navigation performance (according to three metrics from Section~\ref{evaluation_metrics}) with the three baselines discussed in Section~\ref{baselines}.  

\subsubsection{Navigation Performance}
\begin{figure}[!htb]
\centerline{
\includegraphics[width=0.5\textwidth]{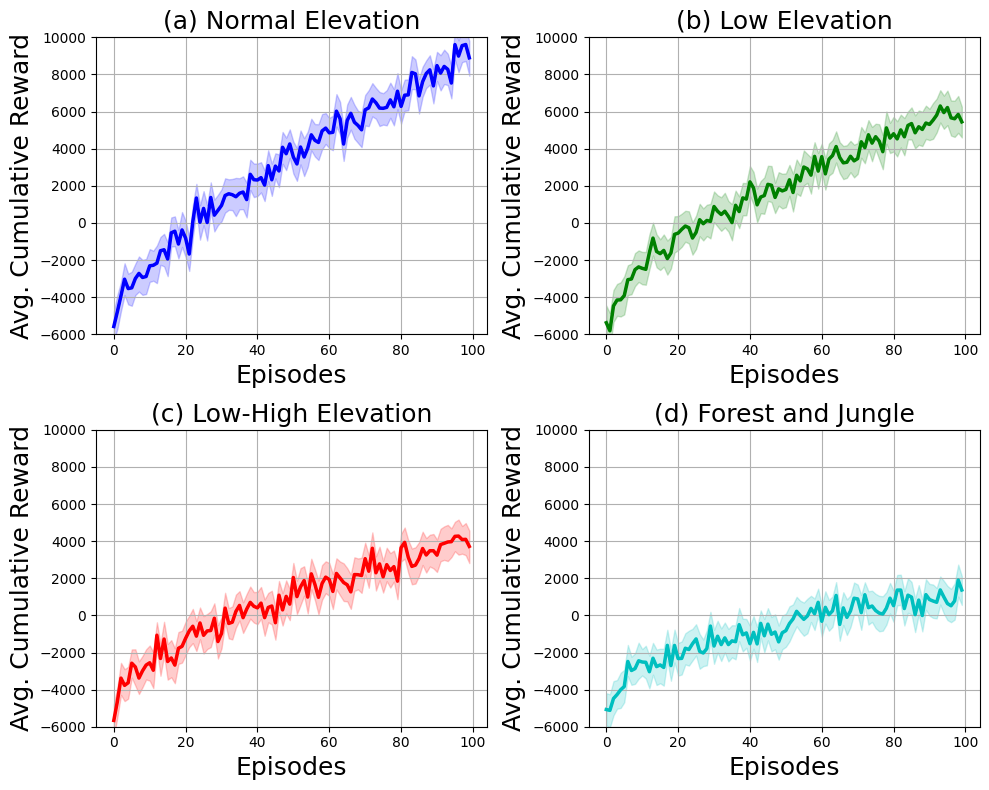}}
    \caption{Performance comparison of \textit{CoverNav} under four scenarios: (a) Normal Elevation, (b) Low Elevation, (c) Low-High Elevation, (d) Forest and Jungle  }
\label{performance_comparison}
\end{figure}

Fig.~\ref{huskyTraining} details the methodical process of the robot attempting to locate a place to stay while using a house as shelter.
It can be observed from Fig.~\ref{performance_comparison} that the agent consistently improves its performance across all scenarios as the number of episodes progresses, demonstrating effective learning. In the normal elevation scenario, the robot demonstrated successful navigation through the environment, efficiently locating a suitable shelter spot without encountering significant challenges. The relatively even terrain allowed for smooth traversal and effective decision-making. In the low elevation scenario, which includes features such as buildings, fences, and other minor impediments, the robot successfully navigated while taking into account these obstacles. It showcased its ability to identify and utilize cover in the presence of such environmental elements. In the low-high elevation scenario, the terrain presented a mix of high and low elevations, posing increased complexity for the robot's navigation. Despite the challenges, the robot was able to adapt and successfully locate shelter locations. However, the learning curve in this scenario was steeper compared to the normal elevation scenario, requiring the robot to refine its strategies for navigating uneven terrains. In the forest and jungle environment, characterized by dense foliage and varied terrain, the robot faced the most challenging conditions. Navigating through the intricate landscape, avoiding obstacles, and identifying suitable shelter spots proved to be challenging tasks. The robot encountered difficulties such as getting stuck or colliding with obstacles in this scenario. However, as the training progressed, the robot gradually improved its navigation skills and adapted to the complex terrain. The learning curve showed a positive trend, indicating the robot's ability to learn and make progress in navigating through the forest environment over time.
 
% As shown in Table~\ref{tab:terrain_comparison}, we compare the performance metrics of the DWA, TERP, and \textit{CoverNav} methods for robot navigation in various terrains.

% From the table, it can be seen that \textit{CoverNav} achieves similar performance in terms of goal-reaching, obstacle avoidance, and uneven terrain. Furthermore, the trajectory length for \textit{CoverNav} and TERP is significantly longer than DWA, indicating that \textit{CoverNav} and TERP are able to find more efficient paths compared to DWA. However, this increase in efficiency comes at a cost of increased execution time. \textit{CoverNav} has the longest execution time among the three methods, which can be attributed to the more complex algorithms used for cover detection and following. However, the difference is not significant, and considering the superior performance of \textit{CoverNav} in Success Rate and Trajectory Length, it can be concluded that \textit{CoverNav} is a promising method for robot navigation in various terrains, especially in scenarios where cover following is important.

\subsubsection{Comparison with Baselines}
Table~\ref{table:performance_comparison} provides a comparison of the performance metrics of the DWA, GA-Nav, TERP, and \textit{CoverNav} methods for robot navigation in various terrains. \textit{CoverNav} demonstrates competitive performance in terms of goal-reaching and obstacle avoidance, with success rates similar to GA-Nav and TERP. 
% Moreover, \textit{CoverNav} and TERP exhibit significantly shorter trajectory lengths compared to DWA, indicating their ability to find more efficient paths.
However, it is worth noting that the execution time for \textit{CoverNav} is longer compared to the other methods, attributed to the more complex algorithms used for cover detection and following. Despite the slightly longer execution time, the superior performance of \textit{CoverNav} in success rate and trajectory length makes it a promising method for robot navigation in various terrains, particularly in scenarios where cover following is of utmost importance

\begin{table}[!htb]
\centering
\resizebox{\columnwidth}{!}{%
\begin{tabular}{@{}llrrrr@{}}
\toprule
Metrics & Method & \begin{tabular}[c]{@{}c@{}}Normal\\ Elevation\end{tabular} & \begin{tabular}[c]{@{}c@{}}Low\\ Elevation\end{tabular} & \begin{tabular}[c]{@{}c@{}}Low-High\\ Elevation\end{tabular} & \begin{tabular}[c]{@{}c@{}}Forest\\ and\\ Jungle\end{tabular} \\ \midrule
\multirow{4}{*}{\begin{tabular}[c]{@{}l@{}}Success\\ Rate (\%)\end{tabular}} & DWA & 72 & 82 & 67 & X \\
 & GA-Nav & 79 & 83 & 71 & X \\
 & TERP & 82 & 83 & 69 & X \\
 & \textbf{CoverNav (Ours)} & \textbf{81} & \textbf{82} & \textbf{70} & \textbf{86} \\\midrule
\multirow{4}{*}{\begin{tabular}[c]{@{}l@{}}Trajectory\\ Length (m)\end{tabular}} & DWA & 5 & 4 & 6 & X \\
 & GA-Nav & 6 & 5 & 7 & X \\
 & TERP & 9 & 5 & 7 & X \\
 & \textbf{CoverNav (Ours)} & \textbf{9} & \textbf{8} & \textbf{7} & \textbf{12} \\\midrule
\multirow{4}{*}{\begin{tabular}[c]{@{}l@{}}Execution\\ Time (s)\end{tabular}} & DWA & 12 & 14 & 17 & X \\
 & GA-Nav & 14 & 13 & 16 & X \\
 & TERP & 10 & 12 & 20 & X \\
 & \textbf{CoverNav (Ours)} & \textbf{15} & \textbf{17} & \textbf{22} & \textbf{25} \\ \bottomrule
\end{tabular}%
}
\caption{Performance Metrics Comparison of DWA, GA-Nav, TERP, and CoverNav (Ours) Methods for Robot Navigation in Various Terrains.}
\label{table:performance_comparison}
\end{table}

% \begin{figure}[htbp]
% \centerline{
% \includegraphics[width=0.5\textwidth]{images/metrics_table.png}}
% \caption*{TABLE I: Performance Metrics Comparison of DWA, TERP,
% and CoverNav Methods for Robot Navigation in Various
% Terrains}
% \label{metricstable}
% \end{figure}

\section{Discussion}
\label{sec:discussion}
% Our results show that our approach is effective in navigating outdoor environments while following covers. The success rate of our approach increases significantly with the number of episodes played. Our approach is also able to successfully navigate through a variety of environments, including normal elevations, low-high elevations, and forest/jungle environments.

In this work, we presented a navigation strategy that effectively navigates outdoor environments while following covers. Our approach demonstrates a significant increase in success rate with the number of episodes played, making it capable of successfully navigating through a variety of environments, including normal elevations, low-high elevations, and forest/jungle environments.

% Moreover, our results show that the performance of our proposed strategy is highly dependent on the length and complexity of the training episodes, as well as the distance of the randomly generated goals. However, when longer-distanced goals are assigned (e.g., 50m), the robot tends to become unstable, making it difficult to achieve the exploration goal. In this scenario, the robot sometimes keeps its collision status even after a new goal has been generated which results in it continuously receiving a negative reward and fails to converge. One approach to addressing this issue could be to adjust the reward function or the training process to provide more guidance to the agent. Additionally, the training process could be modified to gradually increase the difficulty of the goals assigned to the agent, starting with shorter distances and simpler environments and gradually working up to longer distances and more complex environments. This would give the agent time to gradually develop the skills and strategies needed to navigate more challenging environments without becoming unstable.

Furthermore, our analysis highlights the significance of the training episode length, complexity, and the distance of randomly generated goals in determining the performance of our proposed strategy. We observed that when longer-distanced goals, such as those set at 50 meters, were assigned, the robot exhibited instability, making it challenging to accomplish the exploration objective. This instability was evident as the robot continued to collide even after a new goal was generated, leading to a persistent negative reward signal and a failure to converge. To address this issue, several strategies can be employed. Firstly, adjusting the reward function or fine-tuning the training process can provide clearer guidance to the learning agent, encouraging it to prioritize collision avoidance and convergence. By emphasizing the importance of avoiding collisions and rewarding successful navigation, the learning agent can better understand the desired behavior.

In addition, modifying the training process to gradually increase the difficulty of the assigned goals can be beneficial. Starting with shorter distances and simpler environments allows the agent to gradually develop the necessary skills and strategies for navigating challenging environments without succumbing to instability. This progressive approach enables the agent to acquire robust and adaptive navigation capabilities, enhancing its performance in longer-distanced goal scenarios.

It is important to strike a balance between training difficulty and stability to ensure effective learning. By carefully designing the training curriculum, gradually introducing more complex scenarios, and providing adequate training resources, we can foster the development of a resilient and proficient navigation policy that can handle a wide range of goal distances and environmental complexities.

Moreover, the current implementation of the navigation strategy is performed offline, and there is a need for further quantitative validation to assess its real-time performance in unpredictable jungle environments. While evaluations were conducted in emulated environments using Unity, which provides useful insights, it is essential to validate the strategy in real-world scenarios to ensure its effectiveness. A small-scale validation in a real-world setup would provide valuable evidence of its applicability and help bridge the gap between simulated and real-world use cases.

% In successful cover-following navigation, the agent converges when it consistently achieves a high success rate in reaching the goal while following the cover and avoiding obstacles and high elevations. The point at which the learning curve plateaus can give an indication of when the agent has converged to a near-optimal policy. If the learning curve does not plateau but continues to increase, it may suggest that the agent has not yet fully converged and may benefit from further training.

% Regarding the erratic behaviors of the learning curve, we identified that the generation of unreachable goals in the unstructured simulation environment is the primary reason. This causes the robot to enter a vicious cycle of receiving penalties from the reward function for detecting collisions, leading to erratic and unstable learning curves. To mitigate this issue, we could explore the use of more structured simulation environments with pre-defined goals and obstacle configurations. This could potentially result in more stable and consistent learning curves, as the robot would be trained on more realistic scenarios.

\section{Conclusion and Future Work}
\label{sec:conclusion}
% We proposed a DRL-based method to move a robot following maximum cover in an unstructured terrain environment, where the agent seeks for natural shelters such as trees and bushes to hide from potential adversarial agents.  While our method generated longer trajectories compared to some existing approaches, it introduced the unique capability of finding cover, which is valuable in contested environments.

We introduced a novel DRL-based method for guiding a robot in unstructured terrain environments, prioritizing maximum cover to enhance its safety and covert capabilities. Our approach enables the robot to identify and utilize natural shelters such as trees and bushes as hiding spots from potential observer agents. Although our method may result in longer trajectories compared to some existing approaches, it provides the distinct advantage of actively seeking and utilizing cover, which is crucial in contested environments.

% In the future, we hope to improve the process of identifying obstacles and covers by incorporating additional factors such as geometry, and dynamic reward systems. We also plan to leverage RGB images from large-scale datasets to train our model for real off-road unstructured terrain environments, addressing challenges such as finding small objects within grassy areas. Furthermore, we intend to explore the integration of imitation learning and deep reinforcement learning to expedite the early training phase. Utilizing RGB images and robot odometry data from the TartanDrive dataset ~\cite{tartandrive}, a large-scale off-road driving dataset, we aim to train our model to recognize different terrain surfaces and achieve more stable trajectories. By advancing these research directions, we aim to improve the practicality and effectiveness of our navigation method in real-world scenarios, opening up opportunities for its deployment in various applications.

In our future work, we plan to enhance the obstacle and cover detection process by incorporating additional factors such as geometry and dynamic reward systems. Additionally, we intend to leverage RGB images from large-scale datasets to train our model specifically for real off-road unstructured terrain environments. This will help address challenges such as detecting small objects within grassy areas. Furthermore, we intend to investigate the integration of imitation learning and deep reinforcement learning techniques to accelerate the initial training phase. We can expedite the learning process and improve the convergence of our navigation model by utilizing pre-existing expert demonstrations.
To facilitate this research, we will utilize RGB images and robot odometry data from the TartanDrive dataset~\cite{tartandrive}, a comprehensive and large-scale off-road driving dataset. By training our model on this rich dataset, we aim to improve its ability to recognize different terrain surfaces and generate more stable and adaptive trajectories.

By advancing these research directions, we aim to enhance the practicality and effectiveness of our navigation method in real-world scenarios. This will open up opportunities for its deployment in various applications, including autonomous vehicles, and military operations.

% Although our approach tends to generate longer trajectories than some state-of-the-art methods like GANav\cite{ganav} and TERP\cite{terp}, it offers the novelty to find cover, which can be useful in contested environments.

% As we continue our research, one of our goals is to accurately identify the obstacle by include not just a binary variable but also additional factors, such as the distance between and shape of the objects, as well as fixed policy versus dynamic reward systems. An outlook is to include RGB images from existing large-scale datasets for the training process, so that the method can be adopted to real off-road unstructured terrain environments and resolve the issues of finding small objects inside grasses. 

% To get out of the stuck mode, using a fixed policy by utilizing the grid and obstacle information and applying A* search based path finding algorithm could be a viable solution. 
% Moreover, we would like to incorporate imitation learning and deep reinforcement learning to reduce the amount of time spent on early training. Using RGB images and robot odometry data from TartanDrive\cite{tartandrive}, a large-scale off-road driving dataset to train the model to distinguish different terrain surfaces, leading to more stable trajectories, is also a pathway we would like to explore. 

\section*{Acknowledgment}
% The work is funded by the U.S. Army grant \texttt{\#W911NF2120076} and NSF Research Experiences for Undergraduates (REU) grant \texttt{\#CNS-2050999}.
This work has been partially supported by U.S. Army Grant \texttt{\#W911NF2120076}, ONR Grant \texttt{\#N00014-23-1-2119}, NSF CAREER Award \texttt{\#1750936},  NSF REU Site Grant \texttt{\#2050999} and  NSF CNS EAGER Grant \texttt{\#2233879}.

The authors would also like to thank Avijoy Chakma, Zahid Hasan, and Dr. Anuradha Ravi for their constructive feedback on this work,  Arthur C. Schang for setting up the Semantic Segmentation in ARL Unity Environment, and special thanks to Wanying Zhu for conducting the initial experiments.

\vspace{12pt}
\bibliographystyle{unsrt}
\bibliography{bibliography}
\end{document}